\documentclass[journal]{IEEEtran}
\usepackage{amsmath,graphicx,cite,epsfig}
\usepackage{multirow}
\usepackage{booktabs}
\usepackage{float}
\usepackage{xspace}
\usepackage{amssymb}
\usepackage{amsmath}
\usepackage{amsthm}
\usepackage{subfigure}
\usepackage{caption}
\usepackage{threeparttable}
\usepackage{diagbox}
\usepackage{url}
\usepackage{bm}
\usepackage{algorithm}
\usepackage{algorithmic}
\usepackage{bm}
\usepackage{amsfonts}

\newcommand{\ieno}{\textit{i.e.}}
\newcommand{\egno}{\textit{e.g.}}
\usepackage[table,xcdraw]{xcolor}
\ifCLASSINFOpdf

\else

\fi

\begin{document}
%
\title{StyleAM: Perception-Oriented Unsupervised Domain Adaption for Non-reference Image Quality Assessment}
%
%
%

\author{Yiting Lu, Xin Li,~\IEEEmembership{Graduate Student Member, IEEE}, Jianzhao Liu, Zhibo Chen,~\IEEEmembership{Senior~Member,~IEEE}
\thanks{This work was supported in part by NSFC under Grant U1908209, 61632001 and the National Key Research and Development Program of China 2018AAA0101400. (Yiting Lu and Xin Li contributed equally to this work.) (Corresponding author: Zhibo Chen.)}
\thanks{Y. Lu, X. Li, J. Liu and Z. Chen are with the CAS Key Laboratory of Technology in Geo-Spatial Information Processing and Application System, University of Science and Technology of China, Hefei 230027, China (e-mail: luyt31415@mail.ustc.edu.cn; lixin666@mail.ustc.edu.cn; jianzhao@mail.ustc.edu.cn; chenzhibo@ustc.edu.cn).}
}  
\maketitle

\begin{abstract}
Deep neural networks (DNNs) have shown great potential in non-reference image quality assessment (NR-IQA). However, the annotation of NR-IQA is labor-intensive and time-consuming, which severely limits their application especially for authentic images. To relieve the dependence on quality annotation, some works have applied unsupervised domain adaptation (UDA) to NR-IQA.
However, the above methods ignore that the alignment space used in classification is sub-optimal, since the space is not elaborately designed for perception.
To solve this challenge, we propose an effective perception-oriented unsupervised domain adaptation method \textbf{StyleAM} for NR-IQA, which transfers sufficient knowledge from label-rich source domain data to label-free target domain images via \textbf{Style} \textbf{A}lignment and \textbf{M}ixup. 
Specifically, we find a more compact and reliable space \ieno, feature style space for perception-oriented UDA based on an interesting/amazing observation, that the feature style (\ieno, the mean and variance) of the deep layer in DNNs is exactly associated with the quality score in NR-IQA.
Therefore, we propose to align the source and target domains in a more perceptual-oriented space \ieno, the feature style space, to reduce the intervention from other quality-irrelevant feature factors. 
Furthermore, to increase the consistency between quality score and its feature style, we also propose a novel feature augmentation strategy Style Mixup, which mixes the feature styles (\ieno, the mean and variance) before the last layer of DNNs together with mixing their labels. Extensive experimental results on two typical cross-domain settings (\ieno, synthetic to authentic, and multiple distortions to one distortion) have demonstrated the effectiveness of our proposed StyleAM on NR-IQA. 
\end{abstract}

\begin{figure*}
    \centering
    \includegraphics[width=0.95\textwidth]{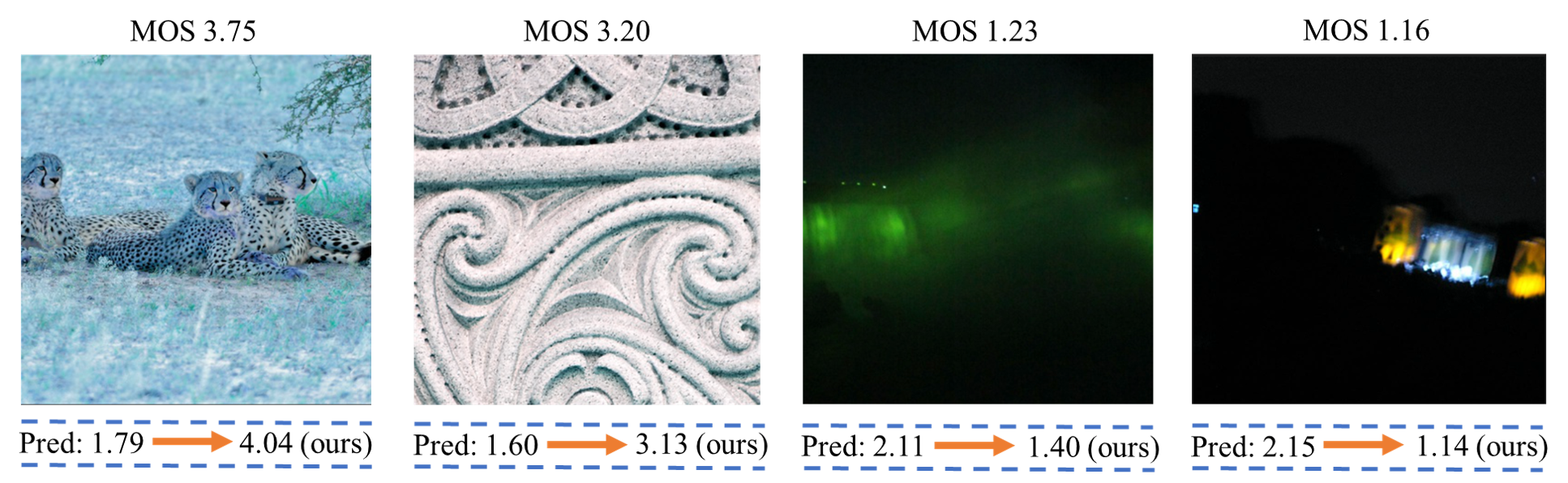}
    \caption{Examples of the domain shift between synthetic dataset and authentic dataset. The IQA metric trained on the synthetic Kadid10k dataset suffers from a severe performance drop when predicting the quality of images from the authetic KonIQ-10K dataset. }  
    \label{fig:teaser}
\end{figure*}

%
\IEEEpeerreviewmaketitle

\section{Introduction}
\label{sec:introduction}
\IEEEPARstart{D}{igital} images are susceptible to various degradations in the process of image acquisition, transmission and processing, which inevitably brings  negative effects on human perception and intelligent visual applications \egno, visual recognition, auto-driving. To address the above problems, image quality assessment (IQA) has been greatly developed to automatically estimate the perceptual quality of images and achieves broad attention. Existing IQA tasks can be roughly divided into three categories, including full-reference, reduced-reference and non-reference, based on whether the reference images are accessible. Among them, non-reference image quality assessment (NR-IQA)
is a more practical and challenging scenario since the lack of reference image.

Traditional NR-IQA algorithms identify the perceptual quality of images by their intrinsic characteristics (\ieno, natural scene statistics (NSS)~\cite{NSS}) in  spatial~\cite{NIQE, BRISQUE} or transform space~\cite{DIIVINE,BLINDS}. However, the poor representation ability of hand-crafted features is far from the human perception system and causes the relative worse results. Thanks to the development of deep learning, 
some pioneering works~\cite{CORNIA,HOSA,CNNIQA,DBCNN,SGDnet,HyperIQA,Rankiqa,zhou2019dual,MEON} utilize well-designed deep neural networks (DNNs) to extract more representative quality-relevant features from images in an end-to-end manner, and achieves excellent performance in many NR-IQA benchmarks~\cite{TID2013,Waterloo,CSIQ,LIVE,Kadid10k,KonIQ,LIVEC}. Nevertheless, learning-based NR-IQA metrics usually require abundant annotated data for training, and easily suffer from poor generalization ability when testing and training images do not meet the independent identical distribution (\ieno, \textit{i.i.d}). A typical example is shown in Fig.~\ref{fig:teaser} that the NR-IQA metric trained with synthetic NR-IQA data failed to identify the perceptual quality of authentic image since the distribution/domain shift between synthetic degradation and authentic degradation. To overcome the domain shift, a na\"ive strategy is to organize professional users to annotate the quality label for images from new domain, while is costly in both time and labor. Moreover, it is impractical to perform annotations traversing over all existed domains especially for authentic degradation.

Thanks to the development of transfer learning, unsupervised domain adaptation (UDA) have been proposed to solve the above challenges, which aims to investigate how to transfer the sufficient knowledge from label-rich source domain data to label-free target domain data under domain shift. The commonly-used UDA methods on high-level vision tasks~\cite{MMD,MMD1,Wa_distance,CORAL,CORNAL,KL,CCD,DANN,UDA_GRL,DANN1,ADDA} focus on learning domain-invariant representation via aligning source and target domains. The alignment strategies can be roughly divided into two categories: 1) aligning the source and target distributions by some metrics~\cite{MMD,MMD1,CORNAL,CORNAL,KL,CCD,Wa_distance} or 2) aligning them by domain adversarial learning~\cite{DANN,UDA_GRL,DANN1,ADDA}. Following the above studies, some works~\cite{UCDA,RankDA} move a step forward and investigate how to transfer the powerful UDA techniques on high-level vision to non-reference image quality assessment (NR-IQA). For instance, UCDA~\cite{UCDA} focuses on the interventions of the distortion diversity and content variation to UDA. Then it divides the target domain into the confident sub-domain and non-confident sub-domain and progressively aligns them in an "easy-to-hard" manner. Chen \textit{et. al}~\cite{RankDA} utilize center loss to learn domain discriminative feature, and then align source and target distribution with maximum mean discrepancy (MMD). However, they ignore the essential fact that the commonly-used alignment strategies in high-level vision tasks are not optimal  for NR-IQA.  

\begin{figure*}
    \centering
    \includegraphics[width=0.9\textwidth]{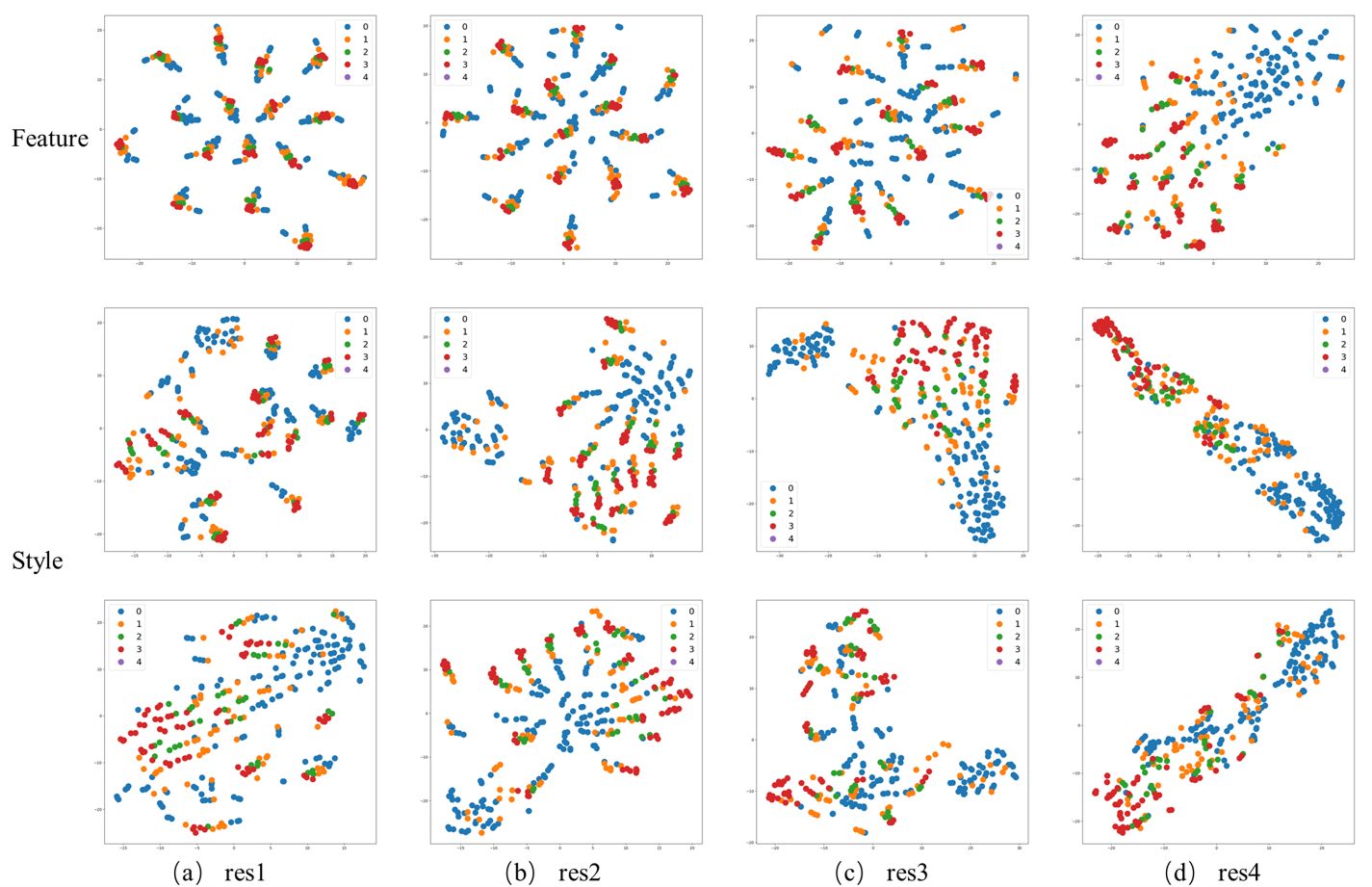}
    \caption{t-SNE visualization of feature and feature style from different resnet layers, where the 2-th row and the 3-th row represent the channel-wise feature mean and channel-wise feature variance respectively, and 'res$x$' means the $x^{th}$ layer in ResNet-18. }
    \label{fig:priliminary}
\end{figure*}

In this paper, we aim to find a more reliable and more perception-oriented space to align the source and target distribution for NR-IQA. In particular, as shown in Fig. ~\ref{fig:priliminary}, we carefully investigated three characteristic of feature captured by ResNet-18 (\ieno, the distributions of feature, channel-wise feature mean and channel-wise feature variance) and analyzed their correlation with quality score. \textbf{Following AdaIN~\cite{Adain}/Mixstyle~\cite{mixstyle}, we call the channel-wise mean and channel-wise variance as feature style.} We surprisingly observed that the style (\ieno, mean and variance) of deep layer in ResNet-18 is more consistent with its quality score than the distribution of features, which reveals that directly aligning two different domains in feature space is not optimal for NR-IQA. Based on the above observation, we propose two novel style-related techniques to better implement the perception-oriented UDA for NR-IQA, respectively as Style Alignment and Style Mixup. Specifically, Style Alignment aims to align the different domains in the feature style (\ieno, mean and variance) space, which can eliminate the intervention of quality-unrelated components existing in the features and thus is a more reliable and perception-oriented space.   Moreover, to increase the consistence between the feature style (\ieno, the mean and variance) and its quality score, we propose a feature augmentation strategy named as Style Mixup, which mixes the feature-wise style (i.e. , the mean and variance) before last layer of DNNs together with mixing their labels. To validate the effectiveness of two techniques, we conducted experiments on two settings, including synthesis datasets to authentic datasets and multiple types of distortions to single type of distortion. The final performance and related ablation studies have shown the superiority of two  techniques for the perception-oriented UDA of NR-IQA.

The contributions of this paper are summarized as follows:
\begin{itemize}
    \item We propose a more effective perception-oriented unsupervised domain adaptation technique \ieno, StyleAM for NR-IQA, which transfers sufficient knowledge from label-rich source data to label-free target data via Style Alignment and Mixup.
    \item Instead of utilizing the commonly-used alignment strategies in UDA, we find a more reliable and perception-oriented alignment space \ieno, feature style space for NR-IQA. Moreover, we carefully design a feature augmentation  \ieno, Style Mixup to increase the consistence between feature style and its quality score. 
    \item Extensive experiments on two typical cross-domain settings \ieno, synthetic distortion to authentic distortion and multiple types of distortions to one type of distortion, have validated the effectiveness of our proposed StyleAM. 
\end{itemize}

\section{Related works}
\label{sec:related work}
\subsection{NR-IQA methods}
Non-reference image quality assessment (NR-IQA) aims to estimate the perceptual quality of images without a clean reference image, which is more practical and challenge problem compared with reference IQA. It can be roughly divided into two categories: distortion-specific and general-purpose approaches. Distortion-specific methods are tailored to specific distortions, such as JPEG compression\cite{JPEGcompressionIQA}, JPEG2000 compression\cite{JPEG2000IQA} and blur\cite{blurIQA}. General-purpose methods are designed based on the assumption that the natural scene statistics (NSS) are highly regular and distortions will break such statistical regularities. Early attempts at general-purpose NR-IQA utilize hand-crafted features based on NSS in spatial domain~\cite{NIQE,BRISQUE,PIQE}, wavelet transform domain~\cite{DIIVINE}, or discrete cosine transform domain~\cite{BLINDS}. Recent years, learning-based methods has significantly advanced the filed of NR-IQA.  CORNIA~\cite{CORNIA} and HOSA~\cite{HOSA} are the early stage of learning-based methods which utilize codebook learning. Later, CNN-based methods\cite{CNNIQA,SGDnet,MEON,zhou2019dual,Rankiqa,lu2022rtn} has further improved the performance of NR-IQA. 

To extend NR-IQA form synthetic distortion to more challenging authentic distortion, which contains hybrid distortions~\cite{li2020learning,liu2020lira}, some works have designed NR-IQA metrics for two kinds of distorted datasets. By utilizing prior authentic distortion knowledge in large natural image classification database, Zhang \textit{et. al}~\cite{DBCNN} pre-train one DNN on large natural image classification database ImageNet~\cite{imagenet} for authentic distortion and another DNN on synthetic distortion dataset, finally fintune the bilinear DNN on synthetic dataset or authentic dataset. Motivated by the diversity of distortion changes and content changes of authentic data, Su \textit{et. al}~\cite{HyperIQA} design the local distortion perception module and hyper network that provide weights for fully-connected (fc) layer to handle the variety of distortion and content in authentic dataset. 
LIQA~\cite{LIQA} focuses on the continual learning scenario in NR-IQA, where the model can sequentially learn new distortions without forgetting the previously learned distortions when the historical training data is inaccessible~\cite{liu2020lira}. However, the above methods are all designed for the ordinary training-testing scenario where the training data and the test data are drawn from the same or similar distribution, ignoring the negative effect of domain shift (\egno, the distribution shift between synthetic distortion dataset and authentic distortion dataset). The NR-IQA model trained on synthetic distortion data has poor performance when directly tested on authentic distortion data as show in Fig.~\ref{fig:teaser}. 

\subsection{Unsupervised Domain Adaptation}
Recently, unsupervised domain adaptation (UDA) has been developed to eliminate the distribution/domain shift~\cite{MMD,MMD1,Wa_distance,CORAL,CORNAL,KL,CCD,DANN,UDA_GRL,DANN1,ADDA,li2021confounder} between source domain and target domain with access to the unlabeled target data. The commonly-used strategy of UDA is to learn domain-invariant feature representation via alignment, which can be roughly divided into divergence-based methods and adversary-based methods. Divergence-based methods try to align the source and target distribution by designing distance measurement functions, including maximum mean discrepancy (MMD)~\cite{MMD1,MMD}, correlation alignment (CORAL)~\cite{CORAL,CORNAL}, contrastive domain discrepancy~\cite{CCD}, kullback-leibler (KL) divergence~\cite{KL},  and wasserstein distance~\cite{Wa_distance}, etc. Among them, MMD~\cite{MMD} can be regard as the weighted sum of  all orders of statistic moment and CORAL~\cite{CORAL} is designed based on the second-order statistical characteristics of the features. Contrastive domain discrepancy~\cite{CCD} 
extended the MMD to the intra-class discrepancy and the inter-class discrepancy by explicitly introducing class information into the metric. KL divergence~\cite{KL} measures the distribution distance between the source and the target domain, and the wasserstein distance~\cite{Wa_distance} calculates the distance between samples from different domains. 
\begin{figure*}
    \centering
    \includegraphics[width=0.9\textwidth]{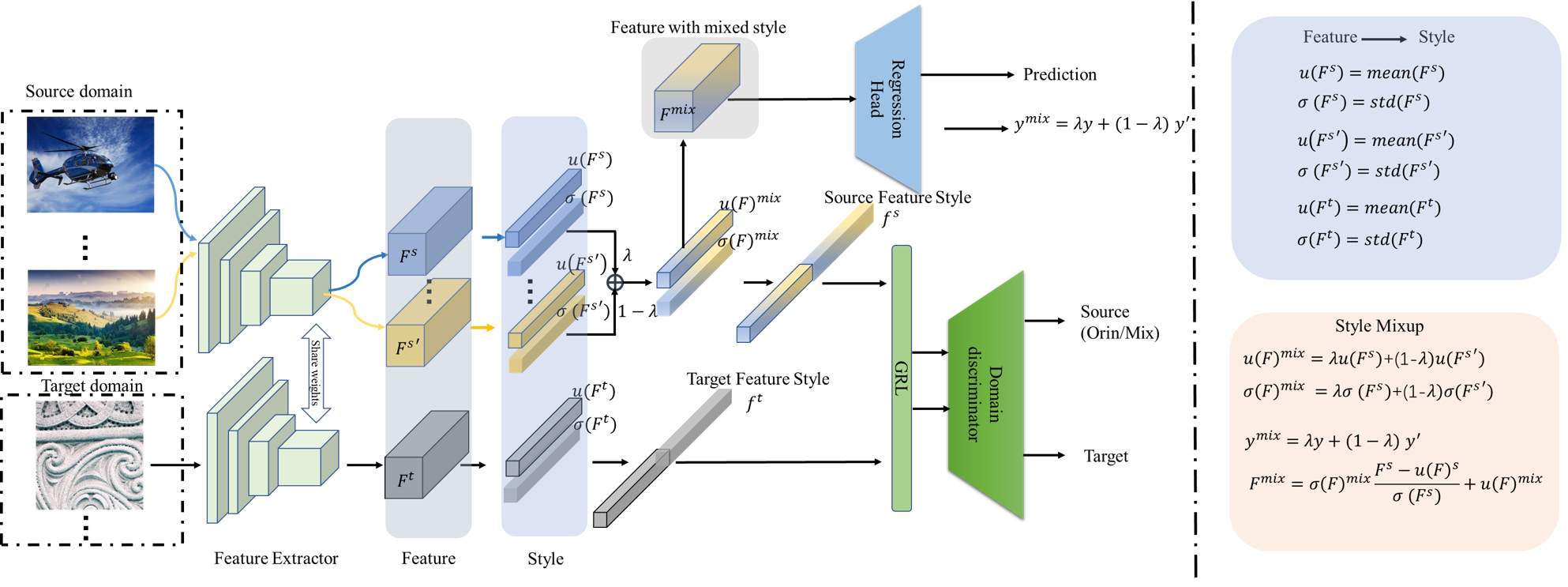}
    \caption{The framework of our proposed StyleAM, containing two bright spots: Style Alignment and Style Mixup. }
    \label{fig:StyleAM}
\end{figure*}

Different from divergence-based methods, adversary-based methods~\cite{DANN,UDA_GRL,DANN1,ADDA} originated from adversarial training~\cite{GAN}, where the domain discriminator aims to distinguish the source and target domains. As an attacker, the feature extractor is devoted to fooling the domain discriminator, thereby learning the domain-invariant feature. However, directly optimizing adversarial loss and classification loss usually cause a sub-optimal solution since the interventions between two losses. To tackle this challenge, wei \textit{et al.}~\cite{metaalign} introduce the meta-learning strategy to remove the interventions between two losses and find a jointly optimal solution.  Despite UDA has achieved great development in the classification task, few works attempt to investigate how to implement UDA in the non-reference image quality assessment (NR-IQA).

As the pioneering UDA works for NR-IQA, UCDA~\cite{UCDA} divided the target domain into the confident and non-confident target subdomains, and then aligned the source and target subdomains with an "easy-to-hard" manner. To transfer knowledge from natural image quality assessment to screen content image quality assessment, Chen \textit{et. al}~\cite{RankDA} utilize the center loss to learn domain discriminative feature, and then align two domains with ranked paired features.
Nevertheless, the above methods ignore a crucial fact that the alignment spaces used in classification task are sub-optimal for NR-IQA.
  Unlike the previous works, in this paper, we aim to find a more reliable and perception-oriented space to align the source and target domains for NR-IQA.

\section{Approach}
In this section, we will clarify the techniques of our StyleAM in detail. As shown in Fig.~\ref{fig:StyleAM}, our StyleAM is composed of two ingenious designs, \ieno, Style Alignment and Style Mixup. Among them, Style Alignment aims to align the source and target domain in a more compact and reliable space, \ieno, feature style space. As an interesting but effective augmentation strategy, Style Mixup mixes the feature styles of different samples and their labels, which aims to increase the consistency and continuity between feature style and quality score. In the section~\ref{sec:pre}, we describe the problem definition of UDA and the commonly-used adversarial domain adaptation strategy. After that, we give the observation and analysis for why select the feature style as the perception-oriented alignment space in section~\ref{sec:observation}. Finally, in section~\ref{sec:style_align} and section~\ref{sec:style_mixup}, we demonstrate our Style Alignment and Style Mixup in detail and analysis their functionalities, respectively.

\subsection{Preliminary}
\label{sec:pre}
For unsupervised domain adaptation (UDA) of NR-IQA, we are given labeled source domain database $D_s=\{(x_i^s,y_i^s)\}_{i=1}^{n_s}$ and unlabeled target domain database $D_t=\{x_j^t\}_{j=1}^{n_t}$, where $x_i^s$ and $x_j^t$ are images from and source and target domains, respectively. $y_i^s$ is the quality score of source data. The $D_s$ and $D_t$ have a severe domain/distribution gap since different degradation types and contents. The purpose of UDA is to eliminate the domain shift between source and target domains and transfer the knowledge from source domain to unlabeled target domain.

\textbf{Adversarial domain adaptation.}
 In this paper, we adopt the popular adversary-based alignment as baseline to implement the perception-oriented UDA of IQA. The baseline is composed of three typical components, \ieno, the feature extractor/generator $G$, domain discriminator $D$ and a regression head $P$. Particularly, $G$ aims to learn domain-invariant feature by fooling the discriminator $D$. $D$ aims to distinguish different domains and minimizes the domain classification loss $\mathcal{L}_D$. Different from classification task, $P$ is used to regress a continual quality score based on the features extracted by $G$ instead of a distribution. To achieve the adversarial training, we follow the DANN~\cite{DANN1} and utilize the gradient reverse layer (GRL) to bridge the $G$ and $D$, which optimizes the $G$ with the gradient of $D$ multiplying a negative value. This manner can achieve a  efficient adversarial training.

\subsection{Observation}
\label{sec:observation}
The commonly-used alignment strategies in UDA are devoted to learning domain-invariant feature representation. However, the quality-related representation is only a subset of the domain-invariant feature representation. Finding a compact quality-related feature can provide a more reliable and perception-oriented space to align the source and target NR-IQA tasks, which can eliminate the intervention from quality-irrelevant components. To find this reliable alignment space, we investigate three characteristics of features (\ieno, the distributions of feature, channel-wise feature mean, and channel-wise feature variance) and analysis the correlation between them and quality score, respectively. \textbf{\textit{Following AdaIN~\cite{Adain}, we call the channel-wise mean and variance of feature as feature style.}}
Specifically, we selected a simple ResNet-18 and a representative NR-IQA dataset Kadid10k~\cite{Kadid10k} to conduct experiments. As shown in Fig.~\ref{fig:priliminary}, we visualize the distributions of features and feature styles (\ieno, the mean and variance) from different layers in ResNet-18 using t-SNE~\cite{tSNE}. From the figure, we can obtain some interesting/amazing findings as follows: 1) Deeper feature and feature style have higher correlations compared with shallow feature and feature style. 2) Feature styles (\ieno, mean and variance) are more correlated to quality score than features and have a better continual characteristic. Based on the above observations, we find a more compact and reliable space for the perception-oriented UDA of NR-IQA, \ieno, the feature style space.
\subsection{Style Alignment}
\label{sec:style_align}
Based on our finding, that feature style space is a more compact and reliable space to align different domains in NR-IQA task, we propose the \textbf{Style Alignment}, which first extracts the feature styles from source and target domains and then align them in a adversarial training manner~\cite{DANN1}. Specifically, given the samples $x_i^s$ and $x_j^t$ from source and target domains, we can obtain their corresponding features $F_i^s \in \mathbb{R}^{C\times H \times W}$ and $F_j^t \in \mathbb{R}^{C\times H \times W}$ with the shared feature extractor $G$ as:
\begin{equation}
 F_i^s, F_j^t = G(x_i^s), G(x_j^t),   
\end{equation}
where $C$, $H$, $W$ are the channel, height and width of features, respectively. Instead of directly aligning the features $F_i^s$ and $F_j^t$, we align the source and target domain in a more perception-oriented space~\ieno, feature style space. Following~\cite{mixstyle, Adain}, we extract the feature style by computing the channel-wise mean $u(F)_c$ and variance $\sigma(F)_c$ of features $F$ as: 
\begin{center}
\begin{equation}
  \begin{split}
      u(F)_c&= \frac{1}{HW}\sum\nolimits_{h=1}^H\sum\nolimits_{w=1}^W F_{c,h,w}\\
      \sigma(F)_c&=\sqrt{\frac{1}{HW}\sum\nolimits_{h=1}^H\sum\nolimits_{w=1}^W (F_{c,h,w}-u(F)_c)^2}\\
  \end{split}
  \label{equ:style}
\end{equation}
\end{center}
Then the feature style of samples $x_i^s$ and $x_j^t$ from source and target domain can be represented as:
\begin{equation}
\centering
      f_i^s,f_j^t=concat(u(F_i^s),\sigma(F_i^s)),concat(u(F_j^t),\sigma(F_j^t)),
  \label{equ:style}
\end{equation}
where $u(F)=\{u(F)_c\}_{c=1}^C$, $\sigma{(F)}=\{ \sigma{(F)_c}\}_{c=1}^C$are the vectors of feature mean and variance.

After obtaining feature style $f_i^s$ and $f_j^t$, domain discriminator $D$ is devoted to determining which is source domain and which belongs to target domain based on feature styles. We can regard it as a two-class classification task and optimize the discriminator $D$ with a binary cross entropy loss:
\begin{equation}
       \mathcal{L}_{D} = - \frac{1}{n_s}\sum\nolimits_{i=1}^{n_s}log(1-D(f_i^s))-\frac{1}{n_t}\sum\nolimits_{j=1}^{n_t}log(D(f_j^t))
   \label{equ:adv_loss}
\end{equation}
With the help of GRL layer used in DANN~\cite{DANN1}, the feature extractor $G$ and discriminator $D$ can be optimized at the same time with loss $\mathcal{L}_D$, which does not require the alternating training used in GAN~\cite{GAN}. Since the purpose of feature extractor $G$ is to fool the discriminator $D$, we can optimize $G$ to the opposite direction of discriminator optimization by changing the sign of the gradient computed with $\mathcal{L_D}$. 

However, the above adversarial  alignment ignores the side effects of partial negative features (\ieno, the partial features that have bad correlation with their quality scores). To identify the negative features and eliminate their effects, we follow the work~\cite{PointDA} and set a threshed $\tau$ for adversarial alignment. When the Spearman rank-order
correlation coefficient (SROCC) of features is lower than $\tau$, it means the features are lower related to the quality score. We relax the alignment by revising the discriminator loss $\mathcal{L}_D$ as:
\begin{equation}
       \mathcal{L}_{D}^{'} = - \frac{1}{n_s}\sum\nolimits_{i=1}^{n_s}log(1-\lvert D(f_i^s)- h\rvert)-\frac{1}{n_t}\sum\nolimits_{j=1}^{n_t}log(D(f_j^t))
   \label{equ:refined adv_loss}
\end{equation}
where $h$ denotes whether the alignment needs to be relaxed. The definition of $h$ can be written as:
\begin{equation}
   \begin{split}
    h &= \begin{cases}
        0,  &\quad if~SROCC(P(F^s),y)>\tau\\
        1,  &\quad  others
            \end{cases}
   \end{split}
   \label{equ:SRCC_thd}
\end{equation}
Meanwhile, the feature extractor $G$ and $P$ are required to learning the knowledge for quality assessment based on labeled source data. In this paper, we utilize $l_2$ loss as the quality score regression loss as:
\begin{equation}
      \mathcal{L}_{q}= \frac{1}{n_s}\sum\nolimits_{i=1}^{n_s}\left\|P(G(x_i^s))-y_i^s\right\|_2,
   \label{equ:qua_loss}
\end{equation}
where $x_i^s$ and $y_i^s$ are the image from source domain and its label.

The overall optimization objectives can be formulated as :
\begin{equation}
    \mathcal{L}_{all}=\mathcal{L}_{q}+\lambda_{adv} \times \mathcal{L}_{D}^{'}.
   \label{equ:loss}
\end{equation}
where $\lambda_{adv}$ is the weight of adversarial loss.

\subsection{Style Mixup}
\label{sec:style_mixup}
To increase the consistency between the feature style and its quality score, we design a feature augmentation strategy \ieno, \textbf{Style Mixup}. There are two related but significantly different augmentation strategies as MixStyle~\cite{mixstyle} and Mixup~\cite{mixup}. The former only mixes the feature style of different samples while keeping their original labels and the latter mixes the images and their labels. Different from them, our \textbf{Style Mixup} mixes the feature styles of different samples and their quality score, which is based on our findings that the deep feature style has a better correlation with the quality score. Specifically, given the feature styles $<u(F^s),\sigma{(F^s)}>$ and $<{u(F^{s^{'}})},\sigma(F^{s^{'}})>$ of the two samples and their quality scores $y^s$ and $y^{s^{'}}$, we set a mixing vector $\lambda$ to mixes the feature styles and labels as:
\begin{equation}
   \begin{split}
    u(F)^{mix}=\lambda u({F^s})+(1-\lambda)u({F^{s^{'}}})\\
    {\sigma{(F)}}^{mix}=\lambda {\sigma(F^s)}+(1-\lambda){\sigma({F^{s^{'}}})}\\
    y^{mix} = \lambda y^s + (1-\lambda)y^{s^{'}} \\
   where~ \lambda \sim{Beta(\alpha,\alpha)},~~~\alpha \in (0,+\infty)\\
   \end{split}
   \label{equ:mixstyle1}
\end{equation}
After obtaining the mixed style $u(F)^{mix}$ and $\sigma{(F)^{mix}}$, we can transfer the features $F^s$ to mixed style like AdaIN/MixStyle~\cite{Adain,mixstyle} as:
\begin{equation}
   \begin{split}
      F^{mix}={\sigma(F)}^{mix} \frac{F^s-u(F^s)}{{\sigma(F^s)}}+{u(F)}^{mix}\\
   \end{split}
   \label{equ:mixstyle}
\end{equation}
As in section 3.3, we concatenate these mixed styles in the channel dimension:
\begin{equation}
   \begin{split}
       f^{mix},f^t=concat(u(F)^{mix},\sigma(F)^{mix}),concat(u(F^t),\sigma(F^t))\\
   \end{split}
   \label{equ:mix feature style concat}
\end{equation}
In this way, we can integrate the \textbf{Style Mixup} into the optimization process of UDA. The final  loss function for our StyleAM is:
\begin{equation}
   \begin{split}
    \mathcal{L}_{all} &= \begin{cases}
        \mathcal{L}_{q}(P(F^{mix}),y^{mix})+\lambda \mathcal{L}_{D}^{'}(f^{mix},f^t),  &\quad if~p>0.5\\
        \mathcal{L}_{q}(P(F^{s}),y)+\lambda \mathcal{L}_{D}^{'}(f^{s},f^t),  &\quad  others
            \end{cases}
   \end{split}
   \label{equ:all loss}
\end{equation}

where $p$ is sampled from uniform distribution limited to [0, 1]. It means we utilize the \textbf{Style Mixup} in a probability of 0.5.
\section{Experiments}

\subsection{Datasets}
\label{sec:dataset}
To verify the effectiveness of our proposed method, we conduct experiments on four IQA datasets, among which two are synthetic datasets (Kadid10k~\cite{Kadid10k}, LIVE~\cite{LIVE}) and the others are authentic datasets (KonIQ-10k~\cite{KonIQ}, LIVEC~\cite{LIVEC}).  

\textbf{Kadid10k.} There are 81 reference images and 10,215 distorted images in the dataset. Each reference image is corrupted by 25 kinds of distortion types (covering 7 categories: blur, color distortion, compression, noise, brightness change, spatial distortion, and sharpness and contrast)  with 5 density levels to obtain 125 distorted images. The  Difference Mean Opinion Score (DMOS) is within the range of $[1~5]$, where higher value of DMOS means higher quality.

\textbf{LIVE.} This dataset contains 29 reference images and 779 distorted images with 6 distortion types, including JPEG, JPEG2000, white noise, Gaussian blur, transmission error in JPEG2000. DMOS value for each distorted image is ranged from 0 to 100, and higher value of DMOS means lower quality.

\textbf{KonIQ-10k.} KonIQ-10k is a large authentic dataset which contains 10,073 distorted images. These images are selected from public multimedia database YFCC100m~\cite{yfcc100m} with about 120 subjective annotations for each distorted image. The mean opinion scores (MOS) value of each distorted image is ranged from 1 to 5, and higher value of MOS means higher quality.

\textbf{LIVEC.} LIVEC contains 1,162 authentic distorted images from different mobile camera devices with more than 350,000 human ratings in total. Each image has an average of 175 subjective annotations. The MOS value of each distorted image is ranged from 0-100, and higher value of MOS means higher quality. 
\begin{table*}[]
\centering
\caption{Performance comparison with SOTA NR-IQA metrics on cross-domain scenario. The average results under four UDA settings are in the last column.}
\label{tab:compare1}
\setlength{\tabcolsep}{3mm}{
\begin{tabular}{c|c|c|c|c|c}
\hline
              & \begin{tabular}[c]{@{}c@{}}kadid10k $\rightarrow$ KonIQ-10k\\ SROCC/PLCC\end{tabular} & \begin{tabular}[c]{@{}c@{}}LIVE$\rightarrow$ KonIQ-10k\\   SROCC/PLCC\end{tabular} & \begin{tabular}[c]{@{}c@{}}Kadid10k$\rightarrow$LIVEC\\ SROCC/PLCC\end{tabular} & \begin{tabular}[c]{@{}c@{}}LIVE$\rightarrow$LIVEC\\  SROCC/PLCC\end{tabular} &
              \begin{tabular}[c]{@{}c@{}}average\\  SROCC/PLCC\end{tabular}\\ \hline
NIQE~\cite{NIQE}          & 0.4469/0.4600                                                                         & 0.4469/0.4600                                                                      &   0.3044/0.3619                                                                                &  0.3044/0.3619
                                 &  0.3756/0.4109\\ \hline
PIQE~\cite{PIQE}          & 0.0843/0.1995                                                                         & 0.0843/0.1995                                                                      &0.2622/0.3617                                                                        &0.2622/0.3617                                                                                    &0.1732/0.2806  \\ \hline
BRISQUE~\cite{BRISQUE}       & 0.1077/0.0991                                                                       &0.037/0.0685                                                                     &  0.2433/0.2512                                                                      &    0.1041/0.1974                                                                    & 0.1230/0.1540 \\ \hline
DBCNN~\cite{DBCNN}         & 0.4126/0.4209                                                                         & 0.5222/0.5171                                                                      & 0.2663/0.2897                                                                     & 0.4554/0.3744                                                                     &0.4141/0.4005    \\ \hline
HyperIQA~\cite{HyperIQA}      & 0.5447/0.5562                                                                         & 0.5911/0.5989                                                                      & 0.4903/0.4872                                                                     & 0.4947/0.4066                                                                     & 0.5302/0.5122  \\ \hline
RankIQA~\cite{Rankiqa}       & 0.6030/0.5511                                                                         & 0.6307/0.5514                                                                      & 0.4906/0.4950                                                                     & 0.5153/0.5363                                                                     & 0.5599/0.5334 \\ \hline
No Adapt      & 0.6346/0.5946                                                                         & 0.5851/0.6115                                                                      & 0.4959/0.5020                                                                     & 0.5793/0.6342                                                                    &  0.5737/0.5855\\ \hline
DANN~\cite{DANN1}          & 0.6382/0.6360                                                                         & 0.6164/0.6512                                                                      & 0.4990/0.4835                                                                     & 0.6044/0.6146                                                                    & 0.5895/0.5963   \\ \hline
UCDA~\cite{UCDA}          & 0.4958/0.5010                                                                         & 0.5830/0.6192                                                                      & 0.3815/0.3584                                                                     & 0.4293/0.4793                                                                      & 0.4724/0.4894  \\ \hline
RankDA~\cite{RankDA}        & 0.6383/0.6227                                                                          &0.6121/0.6417                                                                          & 0.4512/0.4548                                                                      &  0.6319/0.5768                                                                                & 0.5833/0.5740  \\ \hline
StyleAM(ours) & \textbf{0.7002/0.6733}                                                                         & \textbf{0.7351/0.7234}                                                                      & \textbf{0.5844/0.5606}                                                                     & \textbf{0.6292/0.6616}                                                                     &  \textbf{0.6622/0.6547} \\ \hline
\end{tabular}}
\end{table*}
\begin{table*}[]
\centering
\caption{Performance comparison with SOTA NR-IQA metrics on cross distortion scenario of UDA for Kadid10k, where ``X/X” means ``SROCC/PLCC” and type1-7 are denoted as: blur, color distortion, compression, noise, brightness change, spatial distortion, sharpness and contrast. The average results under seven UDA settings are in the last column.}
\label{tab:cross_dis}
\setlength{\tabcolsep}{0.5mm}{
\begin{tabular}{c|c|c|c|c|c|c|c|c}
\hline
                       & others$\rightarrow$type1              & others$\rightarrow$type2   & 
                       others$\rightarrow$type3            & others$\rightarrow$type4      & others$\rightarrow$type5 & 
                       others$\rightarrow$type6 & 
                       others$\rightarrow$type7 &
                       average \\ \hline
NIQE~\cite{NIQE}                   & 0.4263/0.5597  & 0.1080/0.1642  & 0.2460/0.2654   & 0.3027/0.3190   & 0.3187/0.5605    & 0.1424/0.1586    & 0.3102/0.3470  &0.2649/0.3392\\ \hline
PIQE~\cite{PIQE}                   & 0.6785/0.6909  & 0.0987/0.1892 & 0.7113/0.7808   & 0.1836/0.2764    & 0.2945/0.4304 &  0.0248/0.0354 &  0.3695/0.3181 & 0.3372/0.3890\\ \hline
BRISQUE~\cite{BRISQUE}                   &0.0037/0.2171   & 0.2743/0.3488 &  0.1080/0.1313  & 0.0039/0.0663   & 0.2041/0.3577 & 0.0027/0.1121 & 0.1683/0.1318 & 0.1092/0.1950\\ \hline
DBCNN~\cite{DBCNN}                  &0.8218/0.7503   &0.2828/0.2221  & 0.8448/0.8938   & 0.8094/0.7866   &\textbf{0.5900/0.6869}  & 0.4415/\textbf{0.4420} &0.6924/0.7344  &0.6403/0.6451\\ \hline
HyperIQA~\cite{HyperIQA}               &0.5296/0.4981   &0.3016/0.2536  & 0.8667/0.9038   & 0.8247/\textbf{0.8202}   &0.4852/0.6638  & 0.3533/0.4008  & 0.7321/\textbf{0.7839} &0.5847/0.6177 \\ \hline
RankIQA~\cite{Rankiqa}                  & 0.7352/0.7059  &0.3516/0.3642  &0.8168/0.8079    & 0.7849/0.7735   &0.3465/0.5528  & 0.3348/0.4118 & 0.5917/0.6408 & 0.5685/0.6081\\ \hline
No Adapt               & 0.5377/0.5171                          & 0.4440/0.4412                           & 0.8207/0.8567                               & 0.8150/0.8088       & 0.4841/0.5795               & 0.4121/0.4087                 & 0.5274/0.5418       &  0.5773/0.5934             \\ \hline
DANN~\cite{DANN1}                        & 0.4286/0.4045  &0.4127/0.4185  & 0.8706/0.8882   & 0.7524/0.7434   & 0.4556/0.5691 &0.4273/0.4377  & 0.2389/0.2492 & 0.5123/0.5301\\ \hline
UCDA~\cite{UCDA}                      & 0.6043/0.5827  & 0.2885/0.3424 & 0.8430/\textbf{0.9132}  &0.6199/0.6365    & 0.2364/0.3935 & 0.3808/0.4404 &  0.4534/0.5248& 0.4894/0.5480\\ \hline
RankDA~\cite{RankDA}                       & 0.6253/0.5739  &\textbf{0.5690/0.5557}  &0.4106/0.2734    & 0.4695/0.4030   & 0.3158/0.4073 & 0.1958/0.1851 & 0.6765/0.6772 &0.4712/0.4393\\ \hline
      

StyleAM                & \textbf{0.8277/0.8291}                         & 0.4752/0.4719                           & \textbf{0.8763/0.9097 }                              & \textbf{0.8340/0.8199}       & 0.5120/0.6482                      & \textbf{0.4731/0.3628 }                        & \textbf{0.7472/0.7023 }                       &  \textbf{0.6765/0.6777 }  \\ \hline

\end{tabular}}
\end{table*}

\begin{table*}[]
\centering
\caption{Ablation study for three components in StyleAM: Style Mixup, Style Alignment and conditional relaxation (which is in Style Alignment). The average results under four UDA settings are in the last column. }
\label{tab:ablation}
\begin{tabular}{c|c|c|c|c|c}
\hline
                                                                                & \begin{tabular}[c]{@{}c@{}}Kadid10k$\rightarrow$KonIQ-10k\\  SROCC/PLCC\end{tabular} & \begin{tabular}[c]{@{}c@{}}LIVE$\rightarrow$KonIQ-10k\\   SROCC/PLCC\end{tabular} & \begin{tabular}[c]{@{}c@{}}Kadid10k$\rightarrow$LIVEC\\  SROCC/PLCC\end{tabular} & \begin{tabular}[c]{@{}c@{}}LIVE$\rightarrow$LIVEC\\   SROCC/PLCC\end{tabular} &
                                                                                \begin{tabular}[c]{@{}c@{}}average\\  SROCC/PLCC\end{tabular} \\ \hline
No Adapt      & 0.6346/0.5946                                                                         & 0.5851/0.6115                                                                      & 0.4959/0.5020                                                                     & 0.5793/0.6342                                                                    &  0.5737/0.5855\\ \hline
Feature Mixup                                                                           & 0.6217/0.6050~$\downarrow$                                                                        & 0.6380/0.6701 ~$\uparrow$                                                                      & 0.4886/0.4874~$\downarrow$                                                                       & 0.5681/0.5980~$\downarrow$                                                                    & 0.5791/0.5901 ~$\uparrow$ \\ \hline
Style Mixup                                                                     & 0.6711/0.6496 ~$\uparrow$                                                                          & 0.7217/0.7130~$\uparrow$                                                                       & 0.5480/0.5586~$\uparrow$                                                                      & 0.5968/0.6441~$\uparrow$                                                                  &0.6344/ 0.6413 ~$\uparrow$\\ \hline
\begin{tabular}[c]{@{}c@{}}Style Mixup\\ (w.o. label mix)\end{tabular} 
&0.6441/0.6259~$\uparrow$
&0.6772/0.6873~$\uparrow$
&0.5772/0.5651~$\uparrow$
&0.5256/.5583~$\downarrow$
&0.6060/0.6091~$\uparrow$  \\ \hline
Feature Alignment  & 0.6382/0.6360 ~$\uparrow$                                                                         & 0.6164/0.6512 ~$\uparrow$                                                                     & 0.4990/0.4835 ~$\downarrow$                                                                    & 0.6044/0.6146  ~$\uparrow$                                                                    & 0.5895/0.5963~$\uparrow$   \\ \hline
Style Alignment                                                                     & 0.6501/0.6198~$\uparrow$                                                                        & 0.6361/0.6567~$\uparrow$                                                                     & 0.5477/0.4862~$\uparrow$                                                                   
& 0.6055/0.6289~$\uparrow$                                                               
&  0.6098/0.5979~$\uparrow$  \\ \hline
\begin{tabular}[c]{@{}c@{}}StyleAM\\ (w.o. conditional relaxation)\end{tabular}       & 0.6654/0.6526~$\uparrow$                                                                          & 0.6571/0.6557~$\uparrow$                                                                        & 0.5372/0.5362~$\uparrow$                                                                      & 0.5779/0.5980 ~$\downarrow$                                                                   & 0.6094/0.6106~$\uparrow$ \\ \hline
StyleAM                                                                         & 0.7002/0.6733~$\uparrow$                                                                          & 0.7351/0.7234~$\uparrow$                                                                       & 0.5844/0.5606 ~$\uparrow$                                                                     & 0.6292/0.6616~$\uparrow$                                                                   & 0.6622/0.6547~$\uparrow$ \\ \hline
\end{tabular}
\end{table*}

\subsection{Implemented Details}
For all datasets, we linearly rescaled the quality scores to a common range [0-5]. During the training period, we randomly cropped the original image into 384$\times$384, and applied the random horizontally flip for data augmentation. During the testing period, we cropped the original image into 384$\times$384 in the center position. 
We employ a  ResNet-18 (without the final $FC$ layers) pre-trained on ImageNet~\cite{imagenet} as the feature extractor and use two $FC-ReLU$ layers followed by a $Sigmoid$ function to map the $512$-dim latent representation to a scalar quality score.
The domain discriminator $D$ contains two FC layers followed by RELU activation and one FC layer followed by $Sigmoid$ function. 
The details of the hyper-parameters, including the $\alpha$ in beta distribution and the weight of adversarial loss $\lambda_{adv}$, are described in the \textbf{Supplementary}.
All frameworks and experiments are implemented in python using the  Pytorch~\cite{pytorch} library and are trained on a GPU server equipped  with a NIVDIA GeForce 1080Ti.

We first pre-trained the model on the source domain for five epochs based on the Style Mixup, and then utilized Style Alignment with Style Mixup to transfer the knowledge from source domain to target domain. We set the probability of Style Mixup as 0.5, to make it possible to align quality-related features in source style space and mix style space simultaneously. We employed the Adam optimizer with a leaning rate of $1e^{-4}$ and weight decay of $5\times e^{-4}$ to minimize the loss defined in Eq.~\ref{equ:all loss}. During testing, we chose the model obtained at the last training epoch.

We adopted two performance criteria: Spearman
rank-order correlation coefficient (SROCC) and Pearson linear cor-
relation coefficient (PLCC), to measure prediction monotonicity
and precision, respectively.  Higher SROCC/PLCC indicates better correlation between the predicted results and the ground-truth quality scores. 
Before computing PLCC, the predicted quality scores are passed through a non-linear logistic mapping function:
\begin{equation}
    \hat{y}={{\beta }_{1}}(\frac{1}{2}-\frac{1}{\exp ({{\beta }_{2}}(\hat{y}-{{\beta }_{3}}))})+{{\beta }_{4}}\hat{y}+{{\beta }_{5}}
\end{equation}

\subsection{Performance Evaluation}
We utilize the four datasets described in Section \ref{sec:dataset} to construct $2\times2$ ``synthetic $\to$ authentic” cross-domain scenarios and utilize KADID-10K dataset to construct 7 cross-distortion scenarios. We compare the performance of ours against three traditional NR-IQA metrics (NIQE~\cite{NIQE}, BRISQUE~\cite{BRISQUE} and PIQE~\cite{PIQE}), three learning-based methods (RankIQA~\cite{Rankiqa}, DBCNN ~\cite{DBCNN} and HyperIQA~\cite{HyperIQA}), one UDA method adapted from classification task DANN~\cite{DANN}, and two UDA-based methods designed for IQA (UCDA~\cite{UCDA} and RankDA~\cite{RankDA}). Specially, we implemented DANN, UCDA and RankDA with the same backbone ResNet-18 as ours.

\subsubsection{Performance under cross-domain scenario}

Table~\ref{tab:compare1} shows the SROCC/PLCC comparison results of different algorithms under four cross-domain scenarios (Kadid10k $\rightarrow$ KonIQ-10k, LIVE $\rightarrow$ KonIQ-10k, Kadid10k $\rightarrow$ LIVEC and LIVE $\rightarrow$ LIVEC). We can find that our proposed method steadily achieves the best performance compared with other SOTA NR-IQA methods for all cross-domain scenarios.

Firstly, We can observe that the traditional NR-IQA metrics based on NSS and the learning-based NR-IQA metrics designed for authentic distorted data (trained on the source domain) cannot well address the domain shift when directly testing on the target domain. Although HyperIQA~\cite{HyperIQA} takes into account the diversity of content and distortion in authentic distorted data and DBCNN~\cite{DBCNN} adopts a pre-trained model with authentic distortion prior and synthetic distortion prior, they don't consider the domain shift under cross-domain scenario, thus leading performance degradation.


Secondly, compared with the baseline (``No Adapt”), StyleAM has improved the initial performance for all of the four cross-domain scenarios without restriction to dataset size,  which demonstrates the robustness of our method. 


Thirdly, compared with DANN~\cite{DANN1} which directly aligns the high-level features between the source domain and the target domain, our StyleAM which aligns the feature styles (mean and variance) can more effectively mitigate the domain shift, which demonstrates the effectiveness of  our proposed feature style aligning strategy.

Fourthly, compared with UDA methods tailored to IQA (UCDA~\cite{UCDA} and RankDA~\cite{RankDA}), ours can also achieve better performance. In UCDA, it tries to directly align the high-level features of two domains. In RankDA, it aligns the rank feature which reveals  the pairwise quality relationship. However, it selectes the rank pairs according to the inaccurate pseudo scores given by ResNet-18\cite{ResNet} and the final quality scores are generated from inaccurate rank mos, thus it does not bring appparent improvement either.


\subsubsection{Performance under cross-distortion scenario}
Apart from the cross-domain scenarios, we further explore the cross-distortion scenarios within a dataset.
We divide the images in Kadid10k~\cite{Kadid10k} into seven groups according to different distortion categories described in Section~\ref{sec:dataset}.  We conduct leave one distortion type experiments: selecting images of one distortion category as target domain and the remaining images as source domain. For example, images with blur that includes Gaussian blur, Lens blur and Motion blur are selected as the target domain, and then the rest of images can be viewed as the source domain. 
The SROCC/PLCC results of different algorithms are shown in Table~\ref{tab:cross_dis}. From the table, we can see that our proposed StyleAM can improves the baseline performance for all of the seven cross-distortion scenarios. Besides, our method outperforms other IQA metrics by a large margin in most cross-distortion settings except for type3 (compression) and type5 (brightness change). Specially, we find that StyleAM has significant performance improvement on blur distortion category and sharpness (and contrast) distortion category. 

\subsection{Ablation Study}
In the section, we will verify the effectiveness of each key component (including Style Alignment, Style Mixup, and conditional relaxation (one trick used in Style Alignment)) of StyleAM under cross-domain experimental settings. We design six variants of StyleAM. The evaluation results of the six variants together with ``No Adapt” and StyleAM are shown in Table~\ref{tab:ablation}.
\subsubsection{The effect of Style Alignment}
To validate the effectiveness of our proposed Style Alignment, we compare our Style Alignment with the popular Feature Alignment used in recent UDA works. The experimental results are shown in the 5-th and the 6-th rows in Table~\ref{tab:ablation}, respectively. We can find that our Style Alignment outperforms Feature Alignment on all four cross-domain settings (including Kadid10k$\rightarrow$KonIQ-10k, LIVE$\rightarrow$KonIQ-10k, Kadid10k$\rightarrow$LIVEC and LIVE$\rightarrow$LIVEC) by a large margin of 0.02 in average. The above experiments reveals that feature style space is a more reliable and perception-oriented space for alignment compared with feature space.

Another comparison between ``No Adapt” in 1-th row and Style Alignment in Table~\ref{tab:ablation} shows that our Style Alignment based UDA can improve the SROCC about 0.036 in the average result. Combined with Style Mixup, our Style Alignment (\ieno, StyleAM) can further  brings a gain of 0.0885 in terms of SROCC compared with ``No Adapt” in average. Because the Style Mixup can increase the consistency between style and quality score and increase the feature diversity, which improves the transferability of source domain. Moreover, our Style Alignment is robust to different baselines. Based the baseline with Style Mixup in 3-th row, our Style Alignment can achieves the gain of 0.0278 in average, which can demonstrate the superiority of our Style Alignment.

To investigate the effect of conditional relaxation for Style Alignment used in Eq.~\ref{equ:refined adv_loss}, we conduct experiments by removing the conditional relaxation from Eq.~\ref{equ:refined adv_loss} and utilize Eq.~\ref{equ:adv_loss} as the adversarial loss to align the source and target domains. The experimental result is in the 7-th row in Table~\ref{tab:ablation}. We can find that the negative feature leads to a severe performance drop. 
It reveals that the conditional relaxation is necessary for our StyleAM and can eliminate the side effects of negative features to alignment effectively.
\subsubsection{The effect of Style Mixup.}
As a feature augmentation strategy, Style Mixup mixes the feature styles of different samples and their quality scores. There are two similar feature augmentations, respectively as Feature Mixup~\cite{featuremixup} and MixStyle~\cite{mixstyle}. Among them, Feature Mixup mixes the features of different samples and their labels as Eq.~\ref{equ:mixsup}
\begin{equation}
   \begin{split}
    F^{mix}=\lambda F+(1-\lambda)F^{'}\\
    y^{mix}=\lambda y+(1-\lambda)y^{'},\\
   \end{split}
   \label{equ:mixsup}
\end{equation}
where $F$ and $F^{'}$ denote the feature maps of different samples. MixStyle~\cite{mixstyle} only mixes the feature styles of different samples while keeping their original labels, which is equivalent to our Style Mixup without labels mixing. 
To demonstrate the effectiveness of our Style Mixup and clarify the differences of our Style Mixup with the above augmentation strategies, we conduct ablation study for it by three comparisons in Table~\ref{tab:ablation} as follows:

    1) The first comparisons is  shown in the 1-th row and the 3-th row. Based the baseline ``No Adapt”, we directly integrate the Style Mixup into the baseline without Style Alignment. From the table, we can observe that only utilizing Style Mixup can achieves a significant gain of 0.0607 in terms of SROCC. The reason for that is Style Mixup can increase the consistency between feature style and image quality score. Moreover, it can increase the feature diversity, which can improve the generalization ability for new domains.
    
    2) We compare our Style Mixup and Feature Mixup as the second row and third row in Table~\ref{tab:ablation}. We can find directly mixed features and labels achieves almost no performance gain. Because the features contains many quality-irrelevant factors, which intervenes the quality assessment. This is consistent with our observation in Fig.~\ref{fig:priliminary}.
    
    3) We also compare our Style Mixup with MixStyle~\cite{mixstyle} by removing the label mixing in the 4-th row. We can observe that removing the label mixing will cause a performance drop of 0.0284 since it destroy the consistence between feature styles and quality score, which is improper for NR-IQA.

\section{Conclusion}
In this paper, we propose an effective perception-oriented unsupervised domain adaptation method StyleAM for NR-IQA, which is composed of two innovative perception-oriented techniques \ieno, Style Alignment and Style Mixup. Different from the previous UDA works aligning the different domains in the feature space, our StyleAM aligns the source and target domain in a more reliable and perception-oriented space \ieno, style space. Moreover, to increase the consistency between the feature style and quality score, we propose an novel feature augmentation strategy Style Mixup, which mixes the feature styles and quality scores. Based on the obove two style-related techniques, our StyleAM achieves the state-of-the-art performance on two cross domain settings, including synthetic to authentic dataset and multiple distortion to one distortion. Moreover, we believe our interesting/amazing finding, that the feature style space is consistency with quality score, can play an
indispensable role in more aspects of IQA, such as life-long/incremental IQA, full-reference IQA, and domain generalization, etc. We will leave them in the future works. 
\section{Appendix}
Sec.~\ref{sec:exp} describes more experiments on cross-domain scenario. \\
Sec.~\ref{sec:t-SNE} demonstrates the effectiveness of our Style Mixup by t-SNE visualization. \\
Sec.~\ref{sec:hyper} clarifies the details of important hyperparameters in our paper.
\subsection{More experiments on cross-domain scenario}
\label{sec:exp}
To further demonstrate the generalization ability of our StyleAM, we also select the authentic dataset BID as our target domain. BID dataset contains 586 realistic blur images, which are taken by human users in various scenes with different camera apertures and exposition times. As shown in Table~\ref{tab:compare1}, we compare our StyleAM with the state-of-the-art methods on two extra synthetic to authentic experiments, \ieno, Kadid10k $\rightarrow$ BID and LIVE $\rightarrow$ BID, respectively. From the table, our StyleAM can outperform the second best method DANN~\cite{DANN} on Kadid10k $\rightarrow$ BID by 0.05 and exceed the second best method RankDA~\cite{RankDA} by 0.0248, which reveals that our StyleAM owns great generalization ability for different senarios and datasets.
\begin{figure}
    \centering
    \includegraphics[width=0.5\textwidth]{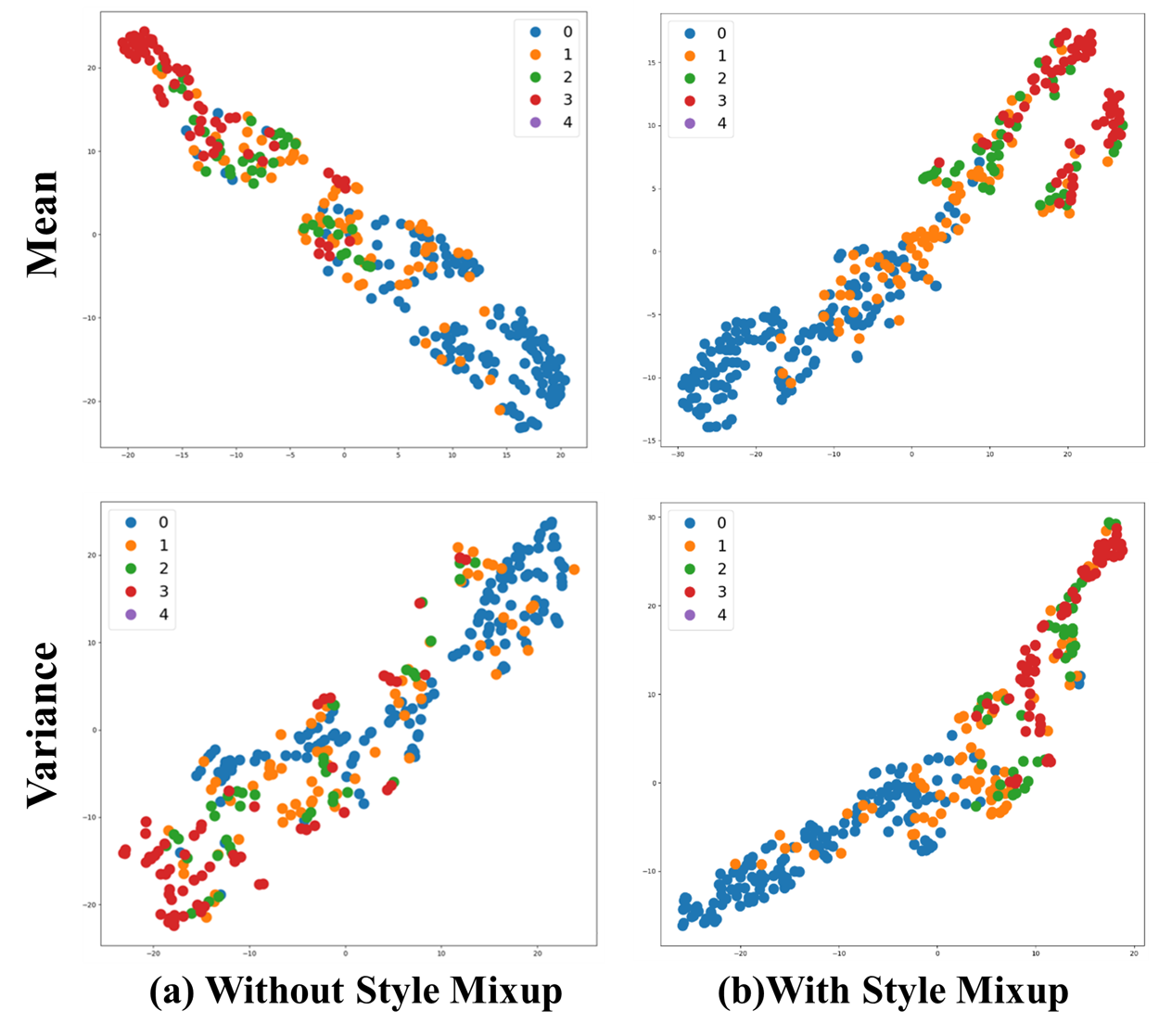}
    \caption{t-SNE visualization of feature style from the last resnet layer before and after using Style Mixup, where the 1-th row and the 2-th row represent the channel-wise feature mean and channel-wise feature variance, respectively. }
    \label{fig:Style Mixup TSNE}
\end{figure}
\begin{table}[htp]
\centering
\caption{Performance comparison with SOTA NR-IQA metrics on cross-domain scenario. The average results under four UDA settings are in the last column.}
\setlength{\tabcolsep}{2mm}{
\begin{tabular}{c|c|c}
\hline
              & \begin{tabular}[c]{@{}c@{}}kadid10k $\rightarrow$ BID \\ SROCC/PLCC\end{tabular} & \begin{tabular}[c]{@{}c@{}}LIVE$\rightarrow$ BID\\   SROCC/PLCC\end{tabular}  \\ \hline
NIQE~\cite{NIQE}          
& 0.3553/0.3812                                                                     
& 0.3553/0.3812\\ \hline
PIQE~\cite{PIQE}        
&  0.2693/0.3506                                                                   
&  0.2693/0.3506 \\ \hline
BRISQUE~\cite{BRISQUE}       
&  0.1745/0.1750                                                                      
&  0.1666/0.1637\\ \hline
DBCNN~\cite{DBCNN}         
& 0.3179/0.2115                                                                         
&0.5273/0.2059  \\ \hline
HyperIQA~\cite{HyperIQA}     
& 0.3794/0.2820                                                                         
& 0.5688/0.5513 \\ \hline
RankIQA~\cite{Rankiqa}       
&0.5101/0.3671                                                                         
&0.6182/0.4166   \\ \hline
No Adapt      
& 0.5600/0.5200
& 0.6996/0.6859\\ \hline
DANN~\cite{DANN1}         
& 0.5861/0.5102                                                                     
& 0.5398/0.4527 \\ \hline
UCDA~\cite{UCDA}          
& 0.3480/0.3907                                                                         
& 0.6532/0.6740 \\ \hline
RankDA~\cite{RankDA}       
&  0.5350/0.5820                                                                      
& 0.7278/0.6911\\ \hline
StyleAM(ours) 
& \textbf{0.6365/0.5669}                                                                        
& \textbf{0.7526/0.7240}\\ \hline
\end{tabular}}
\label{tab:compare1}
\end{table}

\subsection{t-SNE Visualization for Style Mixup}
\label{sec:t-SNE}
As described in our paper, our Style Mixup aims to increase the consistency between the feature style and the quality score. To validate the effectiveness of Style Mixup, we visualize the t-SNE~\cite{tSNE} w.r.t feature styles (\ieno channel-wise feature mean and variance) before and after using Style Mixup, respectively. As shown in Fig.~\ref{fig:Style Mixup TSNE}, with Style Mixup, the feature styles of different samples are more compact and more consistency with their quality labels compared with no Style Mixup, which can further prove the effectiveness of our Style Mixup.
\section{Details of hyperparameters setting}
\label{sec:hyper}
In this section, we will clarify the details of important hyperparamters, including $\alpha$ in the beta distribution of Style Mixup, SROCC threshold $\tau$ used in conditional relaxation, and the weight of adversarial loss $\lambda_{adv}$.
For $\alpha$, we choose the value from the range of $[0.4,0.9]$. Since the distribution of different source domains are different, we set the Srocc threshold $tau$ as 0.9 for the source domain of Kadid10k, and  0.98 for the source domain of LIVE. The $\lambda_{adv}$ is set around 2.

\bibliographystyle{IEEEtran}
\bibliography{references}

\end{document}